\documentclass[10pt,twocolumn,letterpaper]{article}

\usepackage[pagenumbers]{iccv} 

\definecolor{iccvblue}{rgb}{0.21,0.49,0.74}
\usepackage[pagebackref,breaklinks,colorlinks,allcolors=iccvblue]{hyperref}
\usepackage{bm}
\usepackage{multirow}

\title{X$^{2}$-Gaussian: 4D Radiative Gaussian Splatting for Continuous-time Tomographic Reconstruction}

\author{Weihao Yu$^1$ \quad Yuanhao Cai$^2$ \quad Ruyi Zha$^3$ \quad Zhiwen Fan$^4$ \quad \\
Chenxin Li$^1$ \quad Yixuan Yuan$^1$\thanks{Corresponding Author.} \\
\normalsize{$^1$The Chinese University of Hong Kong \quad
$^2$Johns Hopkins University} \\
\normalsize{$^3$The Australian National University \quad 
$^4$University of Texas at Austin}}

\begin{document}
\maketitle
\begin{abstract}
Four-dimensional computed tomography (4D CT) reconstruction is crucial for capturing dynamic anatomical changes but faces inherent limitations from conventional phase-binning workflows. Current methods discretize temporal resolution into fixed phases with respiratory gating devices, introducing motion misalignment and restricting clinical practicality. In this paper, We propose X$^2$-Gaussian, a novel framework that enables continuous-time 4D-CT reconstruction by integrating dynamic radiative Gaussian splatting with self-supervised respiratory motion learning. Our approach models anatomical dynamics through a spatiotemporal encoder-decoder architecture that predicts time-varying Gaussian deformations, eliminating phase discretization. To remove dependency on external gating devices, we introduce a physiology-driven periodic consistency loss that learns patient-specific breathing cycles directly from projections via differentiable optimization. Extensive experiments demonstrate state-of-the-art performance, achieving a 9.93 dB PSNR gain over traditional methods and 2.25 dB improvement against prior Gaussian splatting techniques. By unifying continuous motion modeling with hardware-free period learning, X$^2$-Gaussian advances high-fidelity 4D CT reconstruction for dynamic clinical imaging. Code is publicly available at: \url{https://x2-gaussian.github.io/}.
\end{abstract}    
\section{Introduction}
\label{sec:intro}
Four-dimensional computed tomography (4D CT) has become a cornerstone in dynamic medical imaging\cite{jiang2019augmentation,you2025fb,you2025temporal,jiang2021enhancement}, especially for respiratory motion management in clinical applications such as image-guided radiotherapy (IGRT) \cite{davis2015stereotactic,onishi2011stereotactic}. By capturing both spatial and temporal information of the chest cavity during breathing cycles, 4D CT enables clinicians to monitor and assess respiratory-induced tumor motion and other dynamic anatomical changes during treatment \cite{sonke2005respiratory,fakiris2009stereotactic,baumann2009outcome}. 

Traditional 4D CT reconstruction follows a phase-binning workflow. It first divides the projections into discrete respiratory phases using external gating devices that require direct patient contact, followed by independent reconstruction of each phase to obtain a sequence of 3D volumes. Within this framework, 3D reconstruction methods such as Feldkamp-David-Kress (FDK) algorithm \cite{rodet2004cone}, or total variation minimization \cite{solberg2010enhancement,song2007sparseness}, can be directly applied to 4D CT reconstruction. Due to the limited number of projections available per phase, the reconstructed CT images frequently exhibit significant streak artifacts, which degrade the visibility of fine tissue structures. To address this issue, several researchers \cite{zhi2020high,brehm2013artifact,chen2012motion,li2005motion,rit2011comparative} have proposed methods for extracting patient-specific motion patterns to compensate for respiratory motion across different phases. Meanwhile, other studies \cite{jiang2019augmentation,lahiri2023sparse,hu2022prior,jiang2021enhancement,zhi2021cycn} have explored the use of Convolutional Neural Networks (CNNs) to restore details in artifact-contaminated images. 

Recent advances in Neural Radiance Fields (NeRF) \cite{mildenhall2021nerf} have introduced improved methods for CT reconstruction \cite{cai2024structure,zha2022naf}. These approaches enable high-fidelity 3D reconstruction from sparse views, thereby mitigating the projection undersampling issues caused by phase partitioning. The emergence of 3D Gaussian splatting (3DGS) \cite{kerbl20233d} has further facilitated the development of more efficient and higher-quality methods \cite{cai2024radiative,zha2024r}. Despite these progress, the reconstruction of 4D CT still suffers from two challenges rooted in the traditional phase-binning paradigm. Firstly, previous methods simulate 4D imaging through a series of disjoint 3D reconstructions at predefined phases, failing to model the continuous spatiotemporal evolution of anatomy. This discretization introduces temporal inconsistencies, limits resolution to a few static snapshots per cycle, and produces artifacts when interpolating between phases. Secondly, they heavily relies on external respiratory gating devices, not only introducing additional hardware dependencies and potential measurement errors that can compromise reconstruction accuracy, but also imposing physical constraints and discomfort on patients during the scanning process.

To overcome these limitations, we propose X$^2$-Gaussian, a novel framework that achieves genuine 4D CT reconstruction by directly modeling continuous anatomical motion. Firstly, unlike previous approaches that perform sequential 3D reconstructions, our method introduces a dynamic Gaussian motion model that explicitly captures the continuous deformation of anatomical structures over time by extending radiative Gaussian splatting \cite{zha2024r} into the temporal domain. Specifically, we design a spatiotemporal encoder that projects Gaussian properties onto multi-resolution feature planes, effectively capturing both local anatomical relationships and global motion patterns. The encoded features are then processed by a lightweight multi-head decoder network that predicts deformation parameters for each Gaussian at any queried timestamp, enabling true 4D reconstruction without discrete phase binning. Secondly, we introduce a self-supervised respiratory motion learning method to eliminate the requirement of external gating devices. By leveraging the quasi-periodic nature of respiratory motion, our approach learns to estimate the breathing period directly from the projection data through a novel physiology-driven periodic consistency mechanism that enforces temporal coherence across respiratory cycles. This approach fundamentally differs from traditional phase-based methods by transforming the discrete phase assignments into learnable continuous parameters, enabling our model to automatically discover and adapt to patient-specific breathing patterns.

As shown in \autoref{fig:title}, X$^2$-Gaussian exhibits superior reconstruction performance compared to existing state-of-the-art methods, establishing a new benchmark in 4D CT reconstruction. Our contributions can be summarized as follows:
\begin{itemize}  
\item We present X$^2$-Gaussian, the first method to directly reconstruct time-continuous 4D-CT volumes from projections, which bypasses phase binning entirely, enabling motion analysis at arbitrary temporal resolutions.
\item We extend the static radiative Gaussian splatting into the temporal domain. To our knowledge, this is the first attempt to explore the potential of Gaussian splatting in dynamic tomographic reconstruction.
\item We introduce a novel self-supervised respiratory motion learning module that jointly estimates the respiratory cycle and enforces periodic consistency, eliminating reliance on external gating devices.
\item Extensive experiments demonstrate that our method significantly improves reconstruction quality, reduces streak artifacts, and accurately models respiratory motion, while also showing potential for automatic extraction of various clinical parameters.
\end{itemize}

\begin{figure*}[t]
  \centering
  \includegraphics[width=0.99\linewidth]{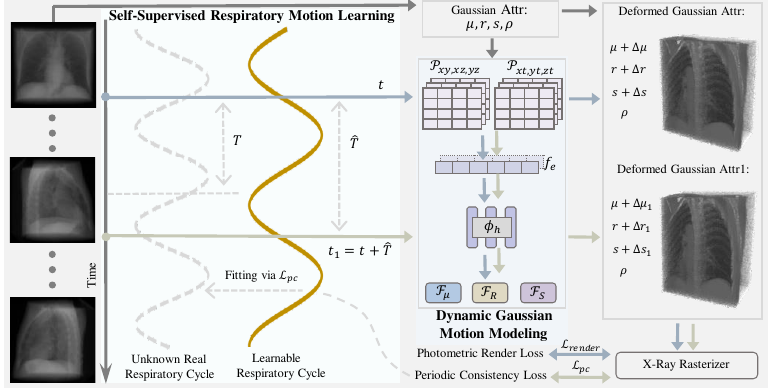}
   \caption{Framework of our X$^2$-Gaussian, which consists of two innovative components: (1) Dynamic Gaussian motion modeling for continuous-time reconstruction; (2) Self-Supervised respiratory motion learning for estimating breathing cycle autonomously.}
   \label{fig:intro}
\end{figure*}

\section{Related Work}
\label{sec:related}

\subsection{CT Reconstruction}
Traditional 3D computed tomography reconstruction methods mainly include two categories: analytical algorithms \cite{feldkamp1984practical,yu2006region} and iterative algorithms \cite{andersen1984simultaneous,sidky2008image,manglos1995transmission,sauer1993local}. Analytical methods estimate the radiodensity by solving Radon transformation and its inverse version. Iterative algorithms are based on optimization over iterations. In recent years, deep learning based models \cite{anirudh2018lose,ghani2018deep,jin2017deep,yu2025toothmaker,lee2023improving,lin2023learning} like CNNs have been employed to learn a brute-force mapping from X-ray projections to CT slices. With the development of 3D deep learning techniques\cite{mildenhall2021nerf,kerbl20233d,yu2022break,yu2019dense,yu2023airwayformer,yu2022tnn,yu2025geot,matsuki2024gaussian,yan2024gs,yugay2023gaussian}, another technical route is to employ the 3D rendering algorithms such as neural radiance fields (NeRF) \cite{mildenhall2021nerf} and 3D Gaussian Splatting (3DGS) \cite{kerbl20233d} to solve the CT reconstruction prolblem in a self-supervised manner, \ie using only 2D X-rays for training. Based on these algorithms, when coping with 4D CTs, researchers typically segment the projections into ten discrete respiratory phases for sequential 3D reconstruction. This approach not only necessitates external devices for phase measurement during scanning but also impedes accurate modeling of the continuous motion of anatomical structures. Concurrent work \cite{fu2025spatiotemporal} also employs dynamic Gaussian splatting. However, they merely establish ten timestamps corresponding to ten phases, thereby maintaining a discrete representation. In contrast, this paper is dedicated to achieving truly continuous-time 4D CT reconstruction.

\subsection{Gaussian Splatting}
3D Gaussian splatting \cite{kerbl20233d} (3DGS) is firstly proposed for view synthesis. It uses millions of 3D Gaussian point clouds to represent scenes or objects. In the past two years, 3DGS has achieved great progress in scene modeling \cite{wu20244d,yang2023real,zhang2024gaussian,yu2024mip}, SLAM \cite{matsuki2024gaussian,yan2024gs,yugay2023gaussian}, 3D Generation \cite{ren2023dreamgaussian4d,xu2024grm}, medical imaging \cite{cai2024radiative,zha2024r}, \emph{etc.} For instance, Cai \emph{et al.} design the first 3DGS-based method, X-GS \cite{cai2024radiative}, for X-ray projection rendering. Later work R$^2$GS \cite{zha2024r} rectifies 3DGS pipeline to enable the direct CT reconstruction. Nonetheless, these algorithms show limitations in reconstructing dynamic CT volumes. Our goal is to cope with this problem.
\section{Preliminaries}
\label{sec:preliminaries}
Radiative Gaussian Splatting \cite{zha2024r} represents 3D CT using a collection of Gaussian kernels $\mathbb{G} = \{G_{i}\}_{i=1}^{K}$, each characterized by its central position $\bm{\mu}_i \in \mathbb{R}^{3}$, covariance matrix $\bm{\Sigma}_{i} \in \mathbb{R}^{3 \times 3}$, and isotropic density $\rho_i$:
\begin{equation}
G_i(\bm{x}|\rho_i, \bm{\mu}_i, \bm{\Sigma}_i) = \rho_i \cdot \exp\left(-\frac{1}{2}(\bm{x} - \bm{\mu}_i)^T\bm{\Sigma}_i^{-1}(\bm{x} - \bm{\mu}_i)\right).
\end{equation}
The covariance matrix can be decomposed as: $\bm{\Sigma}_{i} = \bm{R}_{i}\bm{S}_{i}\bm{S}_{i}^T\bm{R}_{i}^T$, where $\bm{R}_{i} \in \mathbb{R}^{3\times3}$ is the rotation matrix and $\bm{S}_{i} \in \mathbb{R}^{3\times3}$ is the scaling matrix. Then the total density at position $\bm{x}$ is computed as the sum of all contributed Gaussian kernels:
\begin{equation}
\sigma(\bm{x}) = \sum_{i=1}^N G_i(x|\rho_i, \bm{\mu}_i, \bm{\Sigma}_i).
\end{equation}
For 2D image rendering, the attenuation of X-ray through a medium follows the Beer-Lambert Law \cite{kak2001principles}:
\begin{equation}
    I(\bm{r}) = \log I_0 - \log I'(\bm{r}) = \int \sigma(\bm{r}(t))dt,
\end{equation}
where $I_0$ is the initial X-ray intensity, $\bm{r}(t) = \bm{o} + t\bm{d} \in \mathbb{R}^{3}$ represents a ray path, and $\sigma(\bm{x})$ denotes the isotropic density at position $\bm{x} \in \mathbb{R}^{3}$. Thus,  the final pixel value is obtained by integrating the density field along each ray path
\begin{equation}
I_r(\bm{r}) = \sum_{i=1}^N \int G_i(\bm{r}(t)|\rho_i, \bm{\mu}_i, \bm{\Sigma}_i)dt,
\end{equation}
where $I_r(\bm{r})$ is the rendered pixel value.

\section{Methods}
\label{sec:methods} 
\subsection{Overview}
Given a sequence of X-ray projections $\{I_j\}_{j=1}^{N}$ acquired at timestamps $\{t_j\}_{j=1}^{N}$ and view matrices $\{\bm{M}_j\}_{j=1}^{N}$, our goal is to learn a continuous representation of the dynamic CT volume that can be queried at arbitrary timestamps, thereby overcoming the inherent limitations of discrete phase binning. To accomplish this, as shown in \autoref{fig:intro}, our method seamlessly integrates dynamic Gaussian motion modeling with a self-supervised respiratory motion learning scheme into a unified, end-to-end differentiable framework. Specifically, raw Gaussian parameters are initialized from $\{I_j\}_{j=1}^{N}$ and $\{\bm{M}_j\}_{j=1}^{N}$. Given a timestamp $t_j$, dynamic Gaussian motion modeling module predicts the deformation of each parameter, allowing continuous-time reconstruction. Additionally, we model the respiratory cycle as a learnable parameter and sample another timestamp accordingly. Through carefully designed periodic consistency loss, we mine the real breathing period in a self-supervised way.

\subsection{Dynamic Gaussian Motion Modeling}
\label{sec:dynamic}
To achieve continuous 4D CT reconstruction, we introduce a deformation field that models the anatomical dynamics. At the core of our method is a time-dependent deformation field $\mathcal{D}(\bm{\mu}_i, t)$ that predicts the deformation parameters $\Delta G_i$ for each Gaussian at time $t$. The deformed Gaussians $G^{'}_i$ can be computed as:
\begin{equation}
    G^{'}_i = G_i + \Delta G_i = (\bm{\mu}_i + \Delta\bm{\mu}_i, \bm{R}_i + \Delta\bm{R}_i, \bm{S}_i + \Delta\bm{S}_i, \rho_i),
\label{eq:deform}
\end{equation}
where $\Delta\bm{\mu}_i$, $\Delta\bm{R}_i$, and $\Delta\bm{S}_i$ are the deformation offsets for position, rotation, and scaling, respectively. Our deformation field $\mathcal{D}$ is implemented as a composition of two components: $\mathcal{D} = \mathcal{F} \circ \mathcal{E}$, where $\mathcal{E}$ is a spatiotemporal encoder and $\mathcal{F}$ is a deformation-aware decoder.

\paragraph{Decomposed Spatio-Temporal Encoding.}
To encode the spatiotemporal features of Gaussian primitives, a straightforward approach would be to employ neural networks to directly parameterize $\mathcal{E}$. But such a method may lead to low rendering speed and potential overfitting issues, especially given the sparse projection data in 4D CT reconstruction. Inspired by recent advances in dynamic scene reconstruction \cite{fang2022fast,wu20244d,cao2023hexplane}, we adopt a decomposed approach that factorizes the 4D feature space into a set of multi-resolution K-Planes  \cite{fridovich2023k}, which reduces memory requirements while preserving the ability to model complex spatiotemporal patterns in respiratory motion.

Specifically, given a Gaussian center $\bm{\mu} = (x,y,z)$ and timestamp $t$, we project 4D coordinates $\bm{v} = (x,y,z,t)$ onto six orthogonal feature planes: three spatial planes $\mathcal{P}{xy}$, $\mathcal{P}{xz}$, $\mathcal{P}{yz}$ and three temporal planes $\mathcal{P}{xt}$, $\mathcal{P}{yt}$, $\mathcal{P}{zt}$. Each plane $\mathcal{P} \in \mathbb{R}^{d \times lM \times lM}$ stores learnable features of dimension $d$ at multiple resolutions $l \in {1,...,L}$, where $M$ is the basic resolution, enabling simultaneous modeling of fine local motion and global respiratory patterns. The encoded feature $\bm{f}_e$ is computed through bilinear interpolation across multi-resolution planes:
\begin{equation}
f_e = \oplus_{l} \otimes_{(a,b)} \psi \left(\mathcal{P}_{ab}^l(\bm{v}) \right),
\end{equation}
where $\psi$ denotes bilinear interpolation, $\oplus$ represents feature concatenation, $\otimes$ is Hadamard product, and $(a,b) \in \{(x,y),(x,z),(y,z),(x,t),(y,t),(z,t)\}$. Then $\bm{f}_e$ is further merged through a tiny feature fusion network $\phi_h$ (\ie one layer of MLP) as $\bm{f}_h = \phi_h(\bm{f}_e)$.

\begin{figure}[!t]
  \centering
  \includegraphics[width=0.96\linewidth]{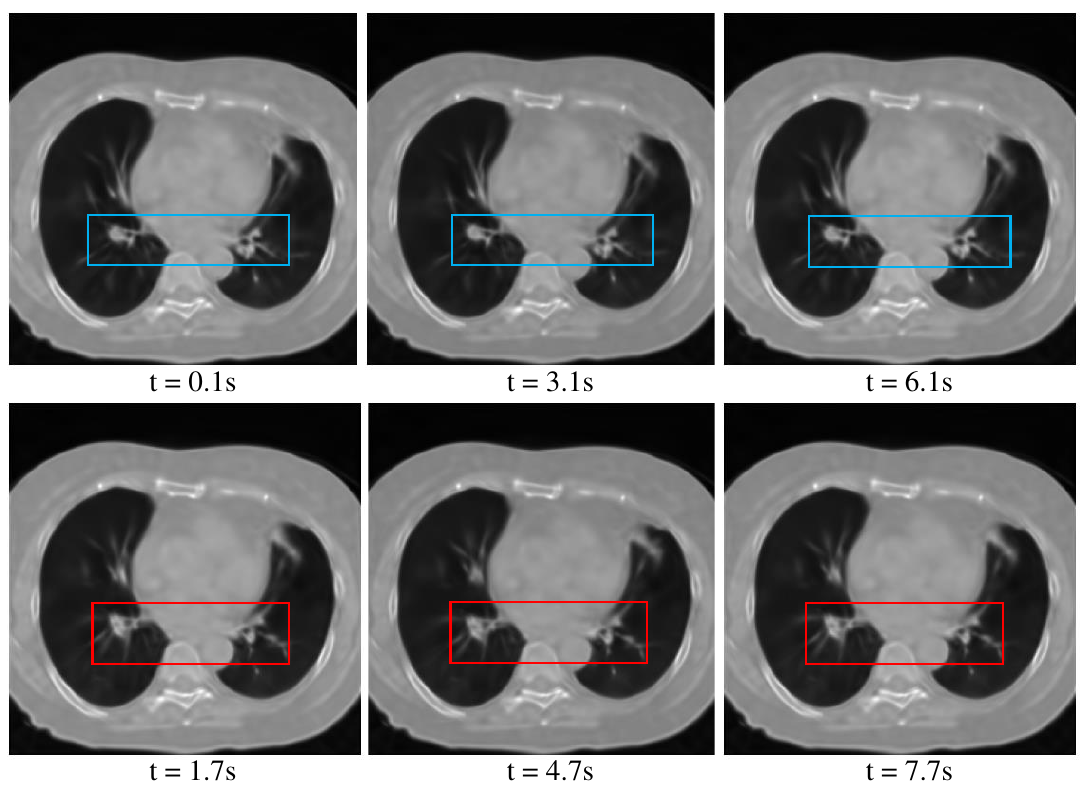}
   \caption{Periodic display of respiratory motion ($T=3s$). A specific anatomical structure (framed by boxes of the same color) at time $t$ has the same position at time $t+nT$.}
   \label{fig:period}
   \vspace{-8pt}
\end{figure}

\paragraph{Deformation-Aware Gaussian Decoding.}
Once the spatiotemporal features are encoded, we employ a lightweight multi-head decoder network $\mathcal{F}$ to predict the deformation parameters for each Gaussian: 
\begin{equation}
\Delta\bm{\mu}, \ \Delta\bm{R}, \  \Delta\bm{S} = \mathcal{F}_{\mu}(\bm{f}_h), \ \mathcal{F}_R(\bm{f}_h), \ \mathcal{F}_S(\bm{f}_h).
\end{equation}
Such decoupled design allows specialized learning of different motion characteristics: position shifts for translational movements, rotation for orientation changes, and scaling for volumetric expansion/contraction. Then the deformed Gaussian parameters at timestamp $t$ can be calculated according to \autoref{eq:deform}. In this way, our dynamic Gaussian motion modeling not only allows independently fine-tune different aspects of motion but also facilitates continuous interpolation across time, yielding smooth temporal transitions in the reconstructed CT volume.

\subsection{Self-Supervised Respiratory Motion Learning}
To eliminate the need for external respiratory gating devices while accurately capturing breathing patterns, we introduce a self-supervised approach that directly learns respiratory motion from projection data. Our method leverages the inherently periodic nature of human respiration to establish temporal coherence across respiratory cycles.

\paragraph{Physiology-Driven Periodic Consistency Loss.}
\label{sec:period}
Respiratory motion exhibits an inherently cyclic pattern, with anatomical structures returning to approximately the same position after each breathing cycle \cite{harris2010speckle}. This physiological characteristic serves as a powerful prior to constrain the reconstruction process. As illustrated in \autoref{fig:period}, a given anatomical position at time $t$ should match its state at time $t + nT$, where $T$ represents the respiratory period and $n$ is an integer. To explicitly encode this periodicity, we enforce a consistency constraint on the rendered images:
\begin{equation}
I(t) = I(t+nT).
\end{equation}
In practice, we define a periodic consistency loss:
\begin{equation}
\mathcal{L}_{pc} = \mathcal{L}_1  \big( I(t), \  I(t+nT) \big) + \lambda_1 \  \mathcal{L}_{ssim} \big( I(t), \ I(t+nT) \big),
\label{eq:learnable_t}
\end{equation}
which encourages the reconstructed images at times $t$ and $t+nT$ to be similar. Here, $\mathcal{L}_1$ and $\mathcal{L}_{ssim}$ are L1 loss and D-SSIM loss \cite{wang2004image}, respectively. This constraint effectively reduces the temporal degrees of freedom in our model by enforcing cyclic coherence, helping to mitigate artifacts and improve reconstruction quality, especially in regions with significant respiratory-induced motion.

\begin{figure}[!t]
  \centering
  \includegraphics[width=0.96\linewidth]{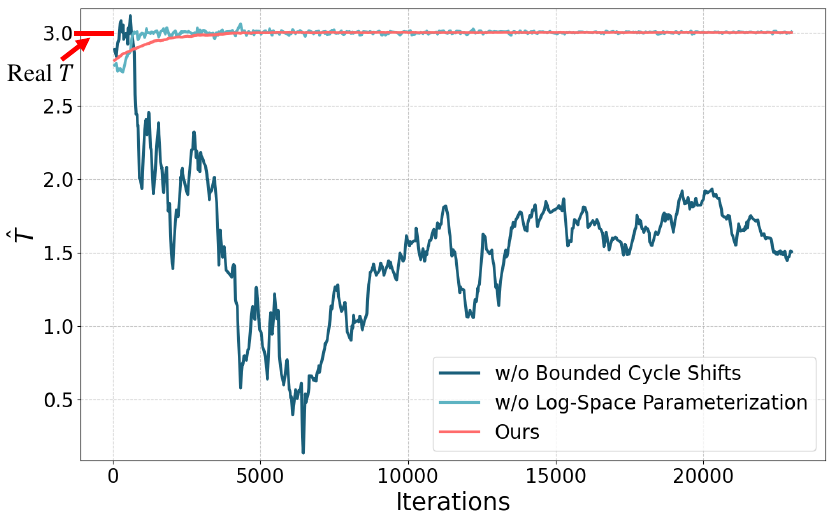}
   \caption{Convergence behavior of the learnable period $\hat{T}$. Without Bounded Cycle Shifts, $\hat{T}$ undergoes wide-ranging oscillations approaching half the true period. Without Log-Space Parameterization, the optimization curve exhibits large oscillations. With both techniques implemented, $\hat{T}$ converges stably and accurately to the correct breathing cycle.}
   \label{fig:learnable_t}
   \vspace{-8pt}
\end{figure} 

\begin{table*}[!t]
  \centering
  \caption{Comparison of our X$^2$-Gaussian with different methods on the DIR dataset.}
     \setlength{\tabcolsep}{1.6mm}{
	\begin{tabular}{ccccccccccccc}
            \toprule
		\multirow{2}{*}{Method}& \multicolumn{2}{c}{Patient1}& \multicolumn{2}{c}{Patient2}& \multicolumn{2}{c}{Patient3}& 
            \multicolumn{2}{c}{Patient4}& \multicolumn{2}{c}{Patient5}& \multicolumn{2}{c}{Average}\\
		  \cmidrule(r){2-3} \cmidrule(r){4-5} \cmidrule(r){6-7} \cmidrule(r){8-9} \cmidrule(r){10-11} \cmidrule(r){12-13} 
		& PSNR& SSIM& PSNR& SSIM& PSNR& SSIM& PSNR& SSIM& PSNR& SSIM& PSNR& SSIM\\
		\hline
            FDK \cite{rodet2004cone}& $34.47$ & $0.836$ & $25.05$ & $0.624$ & $34.23$ & $0.826$& $28.05$ & $0.709$& $25.25$ & $0.638$& $29.41$ & $0.727$\\
            IntraTomo \cite{zang2021intratomo}& $40.04$ & $0.965$& $30.62$ & $0.889$& $33.55$ & $0.888$& $33.00$ & $0.910$& $32.8$ & $0.935$& $34.00$ & $0.917$ \\
            NeRF \cite{mildenhall2021nerf}& $40.85$ & $0.964$ & $32.87$ & $0.917$& $33.43$ & $0.897$& $33.66$ & $0.922$& $34.29$ & $0.955$& $35.02$ & $0.931$\\
            TensoRF \cite{chen2022tensorf}& $33.21$ & $0.907$& $30.32$ & $0.864$& $33.47$ & $0.881$& $33.64$ & $0.813$& $32.40$ & $0.928$& $32.61$ & $0.898$ \\
            NAF \cite{zha2022naf}& $38.21$ & $0.945$& $31.73$ & $0.875$& $34.11$ & $0.900$& $33.95$ & $0.911$& $31.74$ & $0.927$& $33.95$ & $0.912$ \\
            SAX-NeRF \cite{cai2024structure}& $37.21$ & $0.961$& $31.53$ & $0.938$& $36.71$ & $0.929$& $34.30$ & $0.944$& $33.14$ & $0.947$& $34.58$ & $0.942$ \\
            3D-GS \cite{kerbl20233d}& $34.19$ & $0.847$& $22.96$ & $0.713$& $32.53$ & $0.840$& $26.32$ & $0.793$& $29.89$ & $0.812$& $29.18$ & $0.801$ \\
            X-GS \cite{cai2024radiative}& $38.00$ & $0.903$& $25.32$ & $0.739$& $33.54$ & $0.854$& $28.69$ & $0.807$& $28.77$ & $0.793$& $30.86$ & $0.819$ \\
            R$^2$-GS \cite{zha2024r}& $40.51$ & $0.966$& $33.75$ & $0.921$& $39.66$ & $0.956$& $36.45$ & $0.938$& $35.09$ & $0.937$& $37.09$ & $0.943$ \\
            \hline
            Ours & $\bm{44.6}$ & $\bm{0.978}$& $\bm{35.32}$ & $\bm{0.935}$& $\bm{43.22}$ & $\bm{0.972}$& $\bm{37.18}$ & $\bm{0.942}$ & $\bm{36.36}$ & $\bm{0.947}$& $\bm{39.34}$ & $\bm{0.955}$\\
            \bottomrule
  \end{tabular}
  }
  \label{tab:dir}
\end{table*}

\begin{table}[!t]
  \centering
  \caption{Comparison of our X$^2$-Gaussian with different methods on the 4DLung and SPARE datasets.}
     \setlength{\tabcolsep}{2.0mm}{
	\begin{tabular}{ccccc}
            \toprule
		\multirow{2}{*}{Method}& \multicolumn{2}{c}{4DLung}& \multicolumn{2}{c}{SPARE}\\
		  \cmidrule(r){2-3} \cmidrule(r){4-5}
		& PSNR& SSIM& PSNR& SSIM\\
		\hline
            FDK \cite{rodet2004cone}& $27.03$ & $0.611$ & $14.25$ & $0.359$\\
            IntraTomo \cite{zang2021intratomo}& $34.28$ & $0.939$& $27.29$ & $0.871$\\
            TensoRF \cite{chen2022tensorf}& $34.55$ & $0.937$& $26.91$ & $0.857$\\
            NAF \cite{zha2022naf}& $34.94$ & $0.936$& $28.44$ & $0.893$\\
            X-GS \cite{cai2024radiative}& $29.62$ & $0.705$& $18.20$ & $0.442$ \\
            R$^2$-GS \cite{zha2024r}& $37.31$ & $0.952$& $31.12$ & $0.908$ \\
            \hline
            Ours & $\bm{38.61}$ & $\bm{0.957}$& $\bm{32.24}$ & $\bm{0.922}$\\
            \bottomrule
  \end{tabular}
  }
  \label{tab:4dlung_and_spare}
\end{table}

\paragraph{Differentiable Cycle-Length Optimization.}
In realistic scenarios, the true respiratory cycle $T$ is not available a priori. Hence, we treat it as a learnable parameter $\hat{T}$ within our framework. Instead of being provided externally, $\hat{T}$ is optimized directly from the projection data by backpropagating the periodic consistency loss. This allows the network to automatically discover the breathing period in a self-supervised manner. To ensure numerical stability and avoid harmonic artifacts, we implement two critical designs:
\begin{itemize}
    \item \textbf{Bounded Cycle Shifts:} We restrict the integer $n$ in our periodic consistency loss to $n \in \{-1, 1\}$, focusing only on adjacent respiratory cycles. This restriction is critical for avoiding potential ambiguities in period estimation. When using larger values of $n$, the optimization might converge to period estimates that are multiples or divisors of the true period. For example, if the true period $T$ is 3 seconds and our model learns $\hat{T} = 4$ seconds, then with $n = 6$, we would enforce consistency between times $t$ and $t + 24$ seconds, which coincidentally satisfies periodicity (as 24 is divisible by the true period of 3). By limiting $n$ to adjacent cycles, we ensure the model learns the fundamental period rather than its harmonics.
    \item \textbf{Log-Space Parameterization:} We represent $\hat{T} = \exp(\hat{\tau})$ where $\hat{\tau} \in \mathbb{R}$ is an unbounded learnable variable. This ensures positivity and provides smoother gradient updates compared to direct period estimation. This logarithmic parameterization ensures $T$ remains positive, improves numerical stability by preventing extremely small period values, and creates a more uniform gradient landscape for optimization.
\end{itemize}
As shown in \autoref{fig:learnable_t}, these two technical designs are critical for accurate and stable period estimation. Without bounded cycle shifts, the learned period $\hat{T}$ oscillates with large amplitude approaching sub-harmonics (\ie $T/2$) of the true respiratory period, as the periodic consistency loss can be satisfied by most common multiples of sub-harmonics. Direct optimization in linear space leads to pronounced oscillations in the learning trajectory of $\hat{T}$. With both techniques implemented, $\hat{T}$ converges stably and accurately to the correct breathing cycle. In this way, we reformulate \autoref{eq:learnable_t} as
\begin{equation}
\begin{aligned}
    \mathcal{L}_{pc} & = \mathcal{L}_1  \Big( I(t), \  I\big(t+n\exp(\hat{\tau})\big) \Big) \\
    & + \lambda_1 \  \mathcal{L}_{ssim} \Big( I(t), \ I\big(t+n\exp(\hat{\tau})\big) \Big),
\end{aligned}
\label{eq:learnable_t_log}
\end{equation}
where $n \in \{-1, 1\}$. Then the optimal period $T^*$ can be learned via 
\begin{equation}
{\tau}^* = \mathop{\arg\min}_{\hat{\tau}} \mathcal{L}_{pc}, \ \ \ T^* = \exp({\tau}^*).
\end{equation}
Through this self-supervised optimization approach, our model automatically discovers patient-specific breathing patterns directly from projection data without requiring external gating devices, simplifying clinical workflow while improving reconstruction accuracy.

\subsection{Optimization}
\paragraph{Loss Function.}
We optimize our framework by employing a compound loss function. Similar to $\mathcal{L}_{pc}$, we use L1 loss and D-SSIM loss to supervise the rendered X-ray projections as $\mathcal{L}_{render} = \mathcal{L}_1 + \lambda_2 \  \mathcal{L}_{ssim}$. Following \cite{zha2024r}, we integrate a 3D total variation (TV) regularization term \cite{rudin1992nonlinear} $\mathcal{L}_{TV}^{3D}$ to promote spatial homogeneity in the CT volume. We also apply a grid-based TV loss \cite{fridovich2023k,wu20244d,cao2023hexplane} $\mathcal{L}_{TV}^{4D}$ to the multi-resolution k-plane grids used during spatiotemporal encoding. The overall loss function is then defined as:
\begin{equation}
\mathcal{L}_{total} = \mathcal{L}_{render} + \alpha \, \mathcal{L}_{pc} + \beta \, \mathcal{L}_{TV}^{3D} + \gamma \, \mathcal{L}_{TV}^{4D},
\label{eq:total_loss}
\end{equation}
where $\alpha$, $\beta$, and $\gamma$ are weights that control the relative influence of the periodic consistency and regularization terms.

\paragraph{Progressive Training Procedure.} During training, we first train a static 3D radiative Gaussian splatting model \cite{zha2024r} for 5000 iterations. This warm-up phase ensures that the model effectively captures the underlying anatomical structures from the projection data. After the warm-up period, we extend the framework to its full 4D form. The Gaussian parameters, spatiotemporal encoder/decoder, and the learnable respiratory period parameter $\hat{\tau}$ are jointly optimized using the combined loss $\mathcal{L}_{total}$. This progressive training strategy enables the model to build on a robust 3D reconstruction before incorporating temporal dynamics, resulting in stable convergence and high-quality dynamic reconstruction.
\begin{figure*}[!t]
  \centering
  \includegraphics[width=0.99\linewidth]{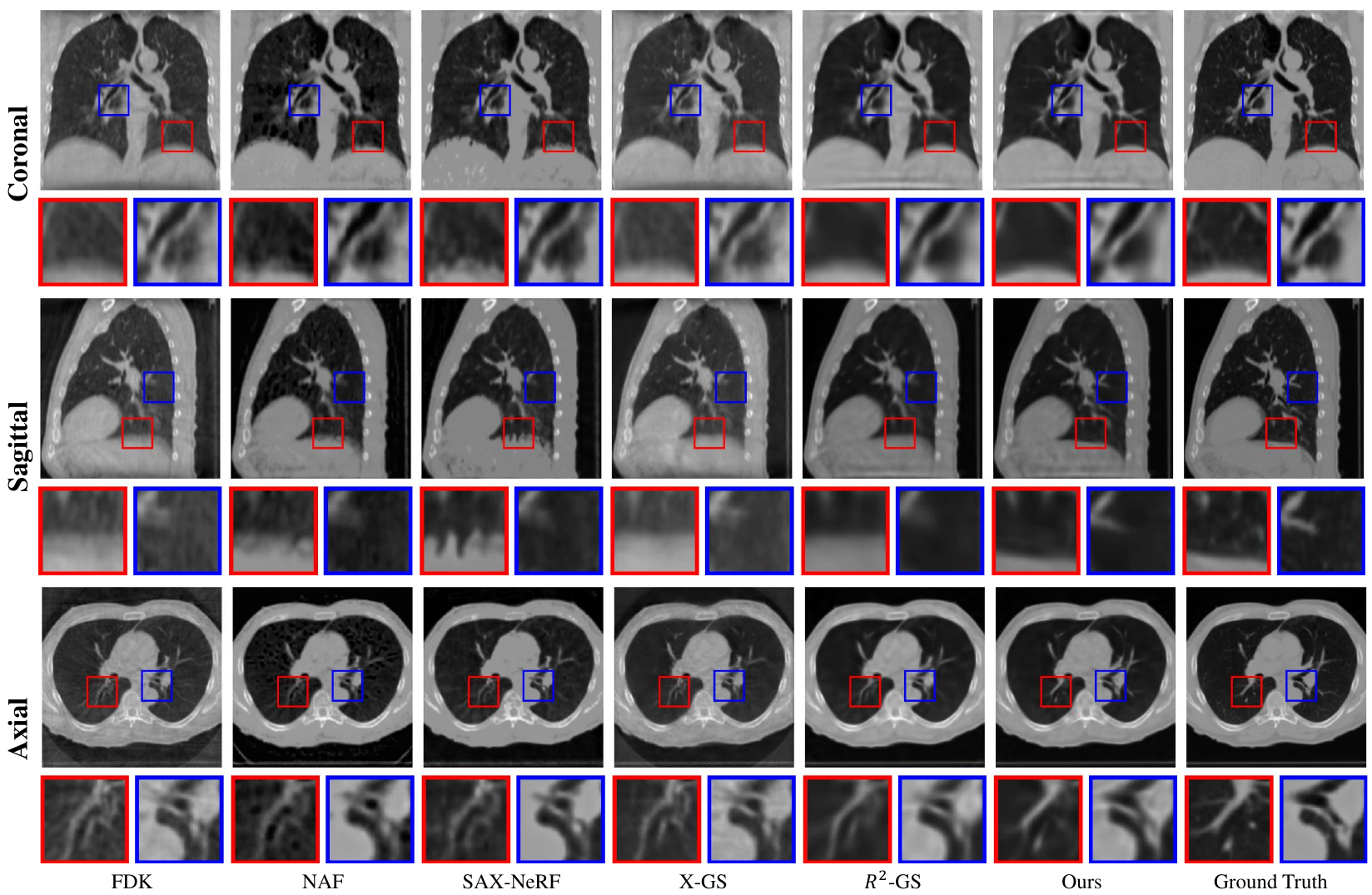}
   \caption{Qualitative comparison of reconstruction results across coronal, sagittal, and axial planes. Our method shows superior performance in modeling dynamic regions (\eg diaphragmatic motion and airway deformation) while preserving finer anatomical details compared to existing approaches.} 
   \label{fig:dir}
   \vspace{-8pt}
\end{figure*}

\section{Experiments}
\label{sec:exp}
\subsection{Dataset and Implementation Details}
We conducted experiments on 4D CT scans from 13 patients across three public datasets: 5 patients from DIR dataset \cite{castillo2009framework}, 5 from 4DLung dataset \cite{hugo2016data}, and 3 from SPARE dataset \cite{shieh2019spare}. Each patient's 4D CT consists of 10 3D CTs from different phases.
We used the tomographic toolbox TIGRE \cite{biguri2016tigre} to simulate clinically significant one-minute 4D CT sampling. The respiratory cycle was configured at 3 seconds, with the corresponding phase determined based on sampling time to obtain X-ray projections. For each patient, 300 projections were sampled, which is substantially fewer than the several thousand projections currently required in clinical settings.

Our X$^2$-Gaussian was implemented by PyTorch \cite{paszke2019pytorch} and CUDA \cite{guide2013cuda} and trained with the Adam optimizer \cite{kingma2014adam} for 30K iterations on an RTX 4090 GPU. Learning rates for position, density, scale, and rotation are initially set at 2e-4, 1e-2, 5e-3, and 1e-3, respectively, and decay exponentially to 10\% of their initial values. The initial learning rates for the spatio-temporal encoder, decoder, and learnable period are set at 2e-3, 2e-4, and 2e-4, respectively, and similarly decay exponentially to 10\% of their initial values. $\hat{\tau}$ was initialized to 1.0296 ($\hat{T} = 2.8$). $\lambda_1$ and $\lambda_2$ in $\mathcal{L}_{pc}$ and $\mathcal{L}_{render}$ were 0.25. $\alpha$, $\beta$, and $\gamma$ in \autoref{eq:total_loss} were set to 1.0, 0.05, and 0.001, respectively. During testing, We used PSNR and SSIM to evaluate the volumetric reconstruction performance. X$^2$-Gausian predicted 10 3D CTs corresponding to the time of each phase, with PSNR calculated on the entire 3D volume and SSIM computed as the average of 2D slices in axial, coronal, and sagittal directions.

\subsection{Results}
\autoref{tab:dir} and \autoref{tab:4dlung_and_spare} illustrate the quantitative results of our X$^2$-Gaussian and SOTA 3D reconstruction methods which follow the phase-bining workflow, including traditional methods (FDK \cite{rodet2004cone}), NeRF-based methods (IntraTomo \cite{zang2021intratomo}, NeRF \cite{mildenhall2021nerf}, TensoRF \cite{chen2022tensorf}, NAF \cite{zha2022naf}, SAX-NeRF \cite{cai2024structure}), and GS-based methods (3D-GS \cite{kerbl20233d}, X-GS \cite{cai2024radiative}, R$^2$-GS \cite{zha2024r}). As can be seen in \autoref{tab:dir}, our method significantly outperforms other approaches in reconstruction quality. Specifically, compared to the traditional FDK method, our approach demonstrates a 9.93 dB improvement in PSNR, achieving approximately a 34\% enhancement. When compared to state-of-the-art methods, our approach surpasses the NeRF-based method SAN-NeRF by 4.76 dB and the GS-based method R2-GS by 2.25 dB. Similar results can be observed in \autoref{tab:4dlung_and_spare}, demonstrating the superiority of our method.

\autoref{fig:dir} shows the quantitative comparison of reconstruction results between our method and existing approaches. Examination of the coronal and sagittal planes shows that our method distinctly captures diaphragmatic motion with remarkable fidelity, which can be attributed to the powerful continuous-time reconstruction capability of X$^2$-Gaussian. Similarly, on the axial plane, X$^2$-Gaussian successfully reconstructs the deformed airways. Additionally, X$^2$-Gaussian preserves fine anatomical details that competing approaches fail to recover, underscoring its effectiveness for high-fidelity volumetric reconstruction.

\begin{figure*}[!t]
  \centering
  \includegraphics[width=0.99\linewidth]{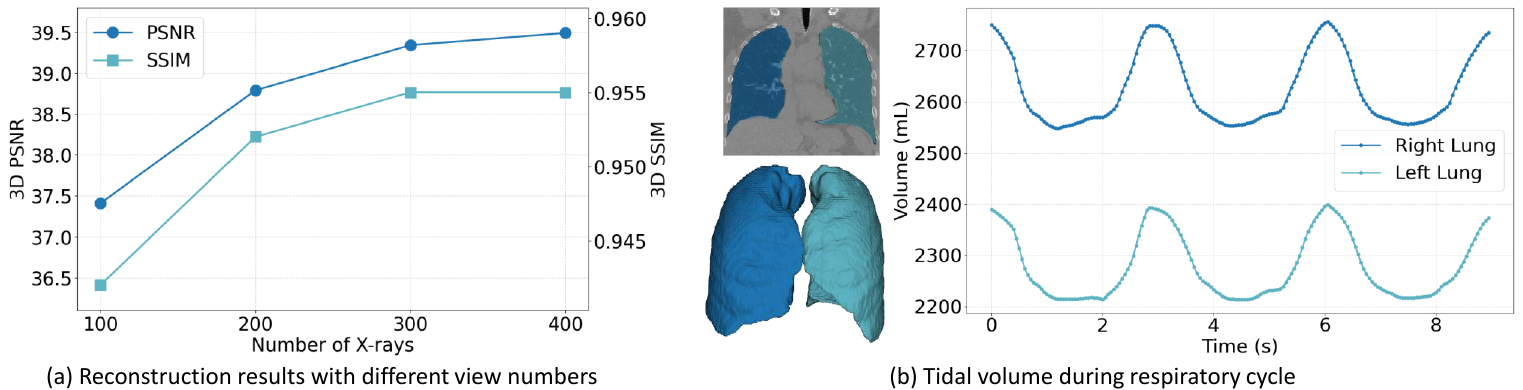}
   \caption{(a) Reconstruction results of X$^2$-Gaussian using different numbers of projections. (b) Temporal variations of lung volume in 4D CT reconstructed by X$^2$-Gaussian. }
   \label{fig:numview}
   \vspace{-10pt}
\end{figure*}

\subsection{Ablation study}
\paragraph{Period Optimization}
\autoref{tab:period} demonstrates the effectiveness of our X$^2$-Gaussian for respiratory cycle estimation and examines how various optimization techniques influence estimation precision. Our approach achieves exceptional accuracy with an average error of just 5.2 milliseconds—approximately one-thousandth of a typical human respiratory cycle. This precision stems from two key technical contributions: Log-Space Parameterization and Bounded Cycle Shifts. Without Log-Space Parameterization, we observe oscillatory convergence behavior that compromises accuracy. More dramatically, when Bounded Cycle Shifts are omitted, the optimization incorrectly converges to harmonic frequencies rather than the fundamental cycle, resulting in a 40-fold increase in estimation error. These findings highlight the critical importance of our optimization framework in achieving reliable respiratory cycle estimation.

\paragraph{Component Analysis}
We conducted ablation experiments on DIR dataset to validate the effect of key components in X$^2$-Gaussian, including the dynamic gaussian motion modeling (DGMM) and self-supervised respiratory motion learning (SSRML). \autoref{tab:ablation} reports the results. As we can see, DGMM extends the static 3D radiative Gaussian splatting model to temporal domain, enabling continuous-time reconstruction and achieving improved 4D reconstruction results. Building upon this foundation, SSRML leverages the periodicity of respiratory motion to directly learn breathing patterns. Remarkably, this approach not only successfully captures specific respiratory cycles but also further enhances reconstruction quality by 0.78 dB, demonstrating its significant contribution to improving temporal coherence and physiological motion plausibility.

\begin{table}[!h]
  \centering
  \caption{Results of respiratory cycle estimation and different optimization techniques used on DIR dataset.}
  \setlength{\tabcolsep}{1.5mm}{
  \begin{tabular}{cccc}
    \toprule
    Method & PSNR& SSIM & Est. error of $T$ (ms)\\
    \hline
    Ours &$\bm{39.34}$ & $\bm{0.955}$ & $\bm{5.2}$ \\
    - Log-sp. param. & $39.32$ & $0.954$ & $12.0$ \\
    - B. cyc. shifts & $39.28$ & $0.954$ &$216.8$ \\
    - Both & $39.23$ & $0.953$ &$914.0$ \\
    \bottomrule
  \end{tabular}
  }
  \label{tab:period}
  \vspace{-10pt}
\end{table}

\paragraph{Hyperparameter Analysis}
We further analyzed the impact of different weights $\alpha$ of periodic consistency loss $\mathcal{L}_{pc}$ in \autoref{tab:ablation}. The optimal performance is achieved when periodic consistency loss and rendering loss are equally weighted (\ie $\alpha$ = 1.0), as this balance enables the model to simultaneously preserve visual fidelity while enforcing physiologically plausible temporal dynamics. When the weighting is either too high or too low, this equilibrium is disrupted, leading to  performance degradation due to either over-constraining the periodic structure at the expense of reconstruction accuracy or prioritizing visual appearance without sufficient temporal coherence.  

\subsection{Discussion}
\paragraph{Projection Numbers}
\autoref{fig:numview} (a) demonstrates the reconstruction results of X$^2$-Gaussian using different numbers of projections. As can be observed, the reconstruction quality gradually improves with an increasing number of available projections. Surprisingly, when compared with \autoref{tab:dir}, we found that even when trained with only 100 X-ray images, our method still achieves better reconstruction results than the current SOTA method R$^2$-GS using 300 X-rays ($37.41$ dB vs. $37.09$ dB). This clearly demonstrates the powerful capability of our approach.

\begin{table}[!t]
  \centering
  \caption{Ablation studies on components and hyperparameters. DGMM denotes dynamic gaussian motion modeling in \autoref{sec:dynamic}, and SSRML is self-supervised respiratory motion learning in \autoref{sec:period}. $\alpha$ is the weight of periodic consistency loss in \autoref{eq:total_loss}.}
  \begin{tabular}{ccc}
    \toprule
    Method & PSNR& SSIM\\
    \hline
     Baseline & $37.09$ & $0.943$\\
     + DGMM & $38.56$ & $0.947$ \\
     + DGMM + SSRML &$\bm{39.34}$ & $\bm{0.955}$\\
    \hline
    $\alpha$ = 0.1 & $38.86$ & $0.952$ \\
    $\alpha$ = 0.5 & $39.14$ & $0.954$ \\
    $\alpha$ = 1.0 & $\bm{39.34}$ & $\bm{0.955}$ \\
    $\alpha$ = 2.0 & $38.41$ & $0.949$ \\
    \bottomrule
  \end{tabular}
  \label{tab:ablation}
  \vspace{-8pt}
\end{table}

\paragraph{Respiratory Motion Quantification}
We densely sampled our X$^2$-Gaussian reconstructed 4D CT within 9 seconds, resulting in 180 3D CT volumes. With automated segmentation algorithm \cite{hofmanninger2020automatic}, we extracted lung masks and calculated the volumetric changes of the lungs over time, as displayed in \autoref{fig:numview} (b). The pulmonary volume dynamics exhibit a periodic sinusoidal pattern, which precisely correlates with the subject's respiratory cycle, demonstrating that our method successfully models respiratory dynamics while achieving truly temporally continuous reconstruction. Furthermore, clinically relevant parameters can be quantitatively extracted from the volume-time curve: Tidal Volume (TV) is 370 ml, Minute Ventilation (MV) is 7.4 L/min, I:E Ratio is 1:1.9, \emph{etc.} These automatically extracted clinical parameters demonstrate the potential of X$^2$-Gaussian in radiomic-feature-guided treatment personalization.

\section{Conclusion}
\label{sec:conclusion}
This paper presents X$^2$-Gaussian, a continuous-time 4D CT reconstruction framework that leverages dynamic radiative Gaussian splatting to capture smooth anatomical motion. Our method bypasses the limitations of phase binning and external gating by integrating dynamic Gaussian motion modeling with a self-supervised respiratory motion module. Experimental results on clinical datasets demonstrate notable improvements in reconstruction fidelity and artifact suppression. This work bridges the gap between discrete-phase reconstruction and true 4D dynamic imaging, offering practical benefits for radiotherapy planning through improved motion analysis and patient comfort.

{
    \small
    \bibliographystyle{ieeenat_fullname}
    \bibliography{main}
}

\clearpage
\setcounter{page}{1}
\maketitlesupplementary

\begin{table*}[!t]
  \centering
  \caption{Comparison of our X$^2$-Gaussian with different methods on the 4DLung dataset.}
   \renewcommand\arraystretch{1.2}
     \setlength{\tabcolsep}{1.6mm}{
	\begin{tabular}{ccccccccccccc}
            \toprule
		\multirow{2}{*}{Method}& \multicolumn{2}{c}{Patient1}& \multicolumn{2}{c}{Patient2}& \multicolumn{2}{c}{Patient3}& 
            \multicolumn{2}{c}{Patient4}& \multicolumn{2}{c}{Patient5}& \multicolumn{2}{c}{Average}\\
		  \cmidrule(r){2-3} \cmidrule(r){4-5} \cmidrule(r){6-7} \cmidrule(r){8-9} \cmidrule(r){10-11} \cmidrule(r){12-13} 
		& PSNR& SSIM& PSNR& SSIM& PSNR& SSIM& PSNR& SSIM& PSNR& SSIM& PSNR& SSIM\\
		\hline
            FDK \cite{rodet2004cone}& $27.36$ & $0.646$ & $22.98$ & $0.410$ & $28.48$ & $0.662$& $28.76$ & $0.654$& $27.59$ & $0.684$& $27.03$ & $0.611$\\
            IntraTomo \cite{zang2021intratomo}& $30.39$ & $0.926$& $35.73$ & $0.930$& $34.99$ & $0.938$& $35.29$ & $0.941$& $35.02$ & $0.960$& $34.28$ & $0.939$ \\
            TensoRF \cite{chen2022tensorf}& $30.42$ & $0.907$& $36.67$ & $0.931$& $34.64$ & $0.933$& $35.14$ & $0.944$& $35.86$ & $0.969$& $34.55$ & $0.937$ \\
            NAF \cite{zha2022naf}& $30.76$ & $0.901$& $37.46$ & $0.932$& $34.69$ & $0.934$& $35.47$ & \bm{$0.947$}& $36.30$ & $0.964$& $34.94$ & $0.936$ \\
            X-GS \cite{cai2024radiative}& $30.62$ & $0.709$& $25.16$ & $0.526$& $31.45$ & $0.722$& $30.88$ & $0.773$& $29.98$ & $0.792$& $29.62$ & $0.705$ \\
            R$^2$-GS \cite{zha2024r}& $33.19$ & $0.918$& $39.22$ & $0.972$& $37.90$ & $0.960$& $37.29$ & $0.939$& $38.96$ & $0.970$& $37.31$ & $0.952$ \\
            \hline
            Ours & \bm{$34.49$} & \bm{$0.929$}& $\bm{40.44}$ & $\bm{0.957}$& $\bm{39.94}$ & $\bm{0.966}$& $\bm{38.10}$ & $0.943$ & $\bm{40.06}$ & $\bm{0.973}$& $\bm{38.61}$ & $\bm{0.957}$\\
            \bottomrule
  \end{tabular}
  }
  \label{tab:4dlung}
\end{table*}

\begin{table*}[!t]
  \centering
  \caption{Comparison of our X$^2$-Gaussian with different methods on the SPARE dataset.}
   \renewcommand\arraystretch{1.2}
     \setlength{\tabcolsep}{1.6mm}{
	\begin{tabular}{ccccccccc}
            \toprule
		\multirow{2}{*}{Method}& \multicolumn{2}{c}{Patient1}& \multicolumn{2}{c}{Patient2}& \multicolumn{2}{c}{Patient3}&  \multicolumn{2}{c}{Average}\\
		  \cmidrule(r){2-3} \cmidrule(r){4-5} \cmidrule(r){6-7} \cmidrule(r){8-9} 
		& PSNR& SSIM& PSNR& SSIM& PSNR& SSIM& PSNR& SSIM\\
		\hline
            FDK \cite{rodet2004cone}& $9.85$ & $0.232$ & $11.85$ & $0.229$ & $21.04$ & $0.616$& $14.25$ & $0.359$ \\
            IntraTomo \cite{zang2021intratomo}& $27.55$ & $0.889$& $27.83$ & $0.864$& $26.48$ & $0.860$&$27.29$ & $0.871$\\
            TensoRF \cite{chen2022tensorf}& $26.88$ & $0.863$& $27.21$ & $0.832$& $26.64$ & $0.877$ &$26.91$ &$0.857$ \\
            NAF \cite{zha2022naf}& $28.67$ & $0.908$& $29.25$ & $0.880$& $27.39$ & $0.892$& $28.44$& $0.893$\\
            X-GS \cite{cai2024radiative}& $14.16$ & $0.328$& $17.37$ & $0.356$& $23.06$ & $0.652$&$18.20$&$0.442$ \\
            R$^2$-GS \cite{zha2024r}& $30.04$ & $0.907$& $32.06$ & $0.901$& $31.26$ & $0.916$& $31.12$ & $0.908$ \\
            \hline
            Ours & $\bm{31.38}$ & $\bm{0.920}$& $\bm{32.47}$ & $\bm{0.907}$& $\bm{32.87}$ & $\bm{0.939}$& $\bm{32.24}$ & $\bm{0.922}$ \\
            \bottomrule
  \end{tabular}
  }
  \label{tab:spare}
  
\end{table*}
\section{Details of Dataset}
\paragraph{DIR Dataset}
We collected 4D CT scans from the DIR dataset \cite{castillo2009framework}, which were acquired from patients with malignant thoracic tumors (esophageal or lung cancer). Each 4D CT was divided into 10 3D CT volumes based on respiratory signals captured by a real-time position management respiratory gating system \cite{keall20044}. For each patient, the CT dimensions are $256 \times 256$ in the x and y axes, while the z-axis dimension varies from $94$ to $112$ slices. The z-axis resolution is $2.5$ $mm$, and the xy-plane resolution ranges between $0.97$ and $1.16$ $mm$. The CT scan coverage encompasses the entire thoracic region and upper abdomen. Following the approach in literature \cite{zha2022naf,cai2024structure}, we preprocessed the original data by normalizing the density values to the range of $[0, 1]$. We simulated the classical one-minute sampling protocol used in clinical settings by uniformly sampling 300 paired time points and angles within a one-minute duration and a $0$ to $360$ angular range. Based on the respiratory phase corresponding to each timestamp, we selected the appropriate 3D CT volume, and then utilized the tomographic imaging toolbox TIGRE \cite{biguri2016tigre} to capture $512 \times 512$ projections.

\paragraph{4DLung Dataset}
4D CTs in 4DLung dataset \cite{hugo2016data} were collected from non-small cell lung cancer patients during their chemoradiotherapy treatment. All scans were respiratory-synchronized into 10 breathing phases. For each patient, the CT scans have dimensions of $512 \times 512$ pixels in the transverse plane, with the number of axial slices varying between $91$ and $135$. The spatial resolution is $0.9766$ to $1.053$ $mm$ in the transverse plane and $3$ $mm$ in the axial direction. Following the same pipeline as DIR dataset, We captured 300 projections with sizes of $1024 \times 1024$.

\paragraph{SPARE Dataset}
The 4D CT images from the SPARE dataset \cite{shieh2019spare} have dimensions of $450 \times 450$ pixels in the transverse plane and $220$ slices in the axial direction, with an isotropic spatial resolution of $1.0$ $mm$ in all directions. Following the same methodology as the DIR dataset, we acquired 300 projections, each with dimensions of $512 \times 512$ pixels.

\section{Implementation details of baseline methods}
We conducted comparison with various 3D reconstruction methods, which were directly applied to 4D reconstruction under the phase binning workflow. Traditional algorithm FDK \cite{rodet2004cone} was implemented using the GPU-accelerated TIGRE toolbox \cite{biguri2016tigre}. We evaluated five SOTA NeRF-based tomography methods: NeRF \cite{mildenhall2021nerf} (using MLP-based volumetric scene representation) ,IntraTomo \cite{zang2021intratomo} (using a large MLP for density field modeling), TensoRF \cite{chen2022tensorf} (utilizing tensor decomposition for efficient scene representation), NAF \cite{zha2022naf} (featuring hash encoding for faster training), and SAX-NeRF \cite{cai2024structure} (employing a line segment-based transformer). The implementations of NAF and SAX-NeRF used their official code with default hyperparameters, while NeRF, IntraTomo, and TensoRF were implemented using code from the NAF repository. All NeRF-based methods were trained for 150,000 iterations. We also evaluated three SOTA 3DGS-based methods: 3DGS \cite{kerbl20233d} (introducing real-time rendering with 3D Gaussians), X-GS \cite{cai2024radiative} (incorporating radiative properties into Gaussian Splatting), and R$^2$-GS \cite{zha2024r} (proposing a tomographic reconstruction approach to Gaussian Splatting). Since 3DGS and X-GS lack the capability for tomographic reconstruction, following \cite{cai2024radiative}, we leveraged their novel view synthesis abilities to generate an additional 100 X-ray images from new viewpoints for each 3D CT. These synthesized views, together with the training data, were used with the FDK algorithm to perform reconstruction. All 3DGS-based methods used their official code with default hyperparameters. All experiments were executed on a single NVIDIA RTX 4090 GPU.

\section{More Quantitative Results}
\autoref{tab:4dlung} and \autoref{tab:spare} present the comparative results for each patient in the 4DLung dataset and DIR dataset, respectively. Our method achieved optimal reconstruction results for nearly all patients across both datasets.

\end{document}